\documentclass[letterpaper, 10 pt, journal,twoside]{ieeetran}  

\IEEEoverridecommandlockouts                              

\usepackage{graphicx}
\usepackage[ruled,vlined]{algorithm2e}
\usepackage{bbold}
\usepackage{mathtools}

\usepackage{float} 
\usepackage{amsmath} 
\usepackage{amssymb}
\usepackage{amsfonts}
\usepackage{amstext}

\usepackage{amsthm}
\usepackage[T1]{fontenc}
\usepackage[symbolgreek,italic]{mathastext}

\DeclareMathOperator*{\argmin}{arg\,min}
\usepackage{bbm}

\usepackage{fancyhdr}
\fancypagestyle{plain}{
  \fancyhf{}
  \fancyhead[C]{Conference on \LaTeX}     
  \fancyfoot[L]{This is a notice}

}
\usepackage{eso-pic}
\usepackage{hyperref}

\usepackage{booktabs}
\usepackage{multirow}
\usepackage{stfloats}

\AddToShipoutPictureBG*{%
  \AtPageLowerLeft{%
    \setlength\unitlength{1in}%
    \hspace*{\dimexpr0.5\paperwidth\relax}
    \makebox(0,0.75)[c]{\parbox{0.7\paperwidth}{\raggedright\footnotesize© 2021 IEEE.  Personal use of this material is permitted. Permission from IEEE must be obtained for all other uses, in any current or future media, including reprinting/republishing this material for advertising or promotional purposes, creating new collective works, for resale or redistribution to servers or lists, or reuse of any copyrighted component of this work in other works.}}%
}}

\title{Corrective Shared Autonomy for Addressing Task Variability}

\author{Michael Hagenow,$^{1}$ Emmanuel Senft,$^{2}$ Robert Radwin,$^{3}$ Michael Gleicher,$^{2}$ Bilge Mutlu, and $^{2}$ Michael Zinn$^{1}$
\thanks{Manuscript received: October, 15, 2020; Revised December, 30, 2020; Accepted February, 16, 2021.}
\thanks{This paper was recommended for publication by Editor Gentiane Venture upon evaluation of the Associate Editor and Reviewers' comments.
This work was supported by a NASA University Leadership Initiative (ULI) grant awarded to the UW-Madison and The Boeing Company (Cooperative Agreement \# 80NSSC19M0124).}
\thanks{$^{1}$Michael Hagenow and Michael Zinn are with the Department of Mechanical
Engineering, University of Wisconsin--Madison, Madison 53706, USA
        {\tt\small [mhagenow|mzinn]@wisc.edu}}%
\thanks{$^{2}$Emmanuel Senft, Michael Gleicher, and Bilge Mutlu are with the Department of Computer
Sciences, University of Wisconsin--Madison, Madison 53706, USA
        {\tt\small [esenft|gleicher|bilge]@cs.wisc.edu}}
\thanks{$^{3}$Robert Radwin is with the Department of Industrial and Systems Engineering, University of Wisconsin--Madison, Madison 53706, USA
        {\tt\small rradwin@wisc.edu}}
\thanks{Digital Object Identifier (DOI): \href{https://doi.org/10.1109/LRA.2021.3064500}{10.1109/LRA.2021.3064500}.}}%
\begin{document}
\maketitle

\begin{abstract}
Many tasks, particularly those involving interaction with the environment, are characterized by high variability, making robotic autonomy difficult. One flexible solution is to introduce the input of a human with superior experience and cognitive abilities as part of a shared autonomy policy. However, current methods for shared autonomy are not designed to address the wide range of necessary corrections (e.g., positions, forces, execution rate, etc.) that the user may need to provide to address task variability. In this paper, we present \emph{corrective shared autonomy}, where users provide corrections to key robot state variables on top of an otherwise autonomous task model. We provide an instantiation of this shared autonomy paradigm and demonstrate its viability and benefits such as low user effort and physical demand via a system-level user study on three tasks involving variability situated in aircraft manufacturing.
\end{abstract}

\section{INTRODUCTION}
\IEEEPARstart{T}{here} are many instances where small errors can make or break whether an agent fails to complete a task. For example, misalignment or improper torque when installing a screw may cause the screw to strip; too much force can cause a sander to scorch a surface; and small errors in orientation during ultrasound imaging can produce unusable results. A root cause of these errors is the inherent variability of the tasks (e.g., small differences in the screws, different wood properties in each piece, and differences in person-to-person anatomy). This variability makes robotic autonomy challenging in such circumstances as it requires appropriate sensing, modeling, and conditioning for variability to assure that the robot has a sufficiently robust policy for deployment.

We consider an alternate approach to full autonomy where we instead leverage the experience and cognitive abilities of a human collaborator as a part of a shared autonomy system. The premise of this approach is to offload the majority of the execution to the robot and focus the human input on providing sporadic corrections to address small errors in real time. As the initial examples indicate, such a method needs to provide corrections across a spectrum of variables (e.g., position, forces, orientation, execution rate, etc.), which are dictated by the specific task. To address this requirement, we enable users to provide corrections to a subspace of the robot state variables consistent with the needs of a given task. In this work, we propose and implement a shared autonomy method where users provide real-time corrections to a nominal autonomous robot policy, specifically to the robot state variables that are required to resolve issues.

\begin{figure}[t]
\centering
\includegraphics[width=3.40in]{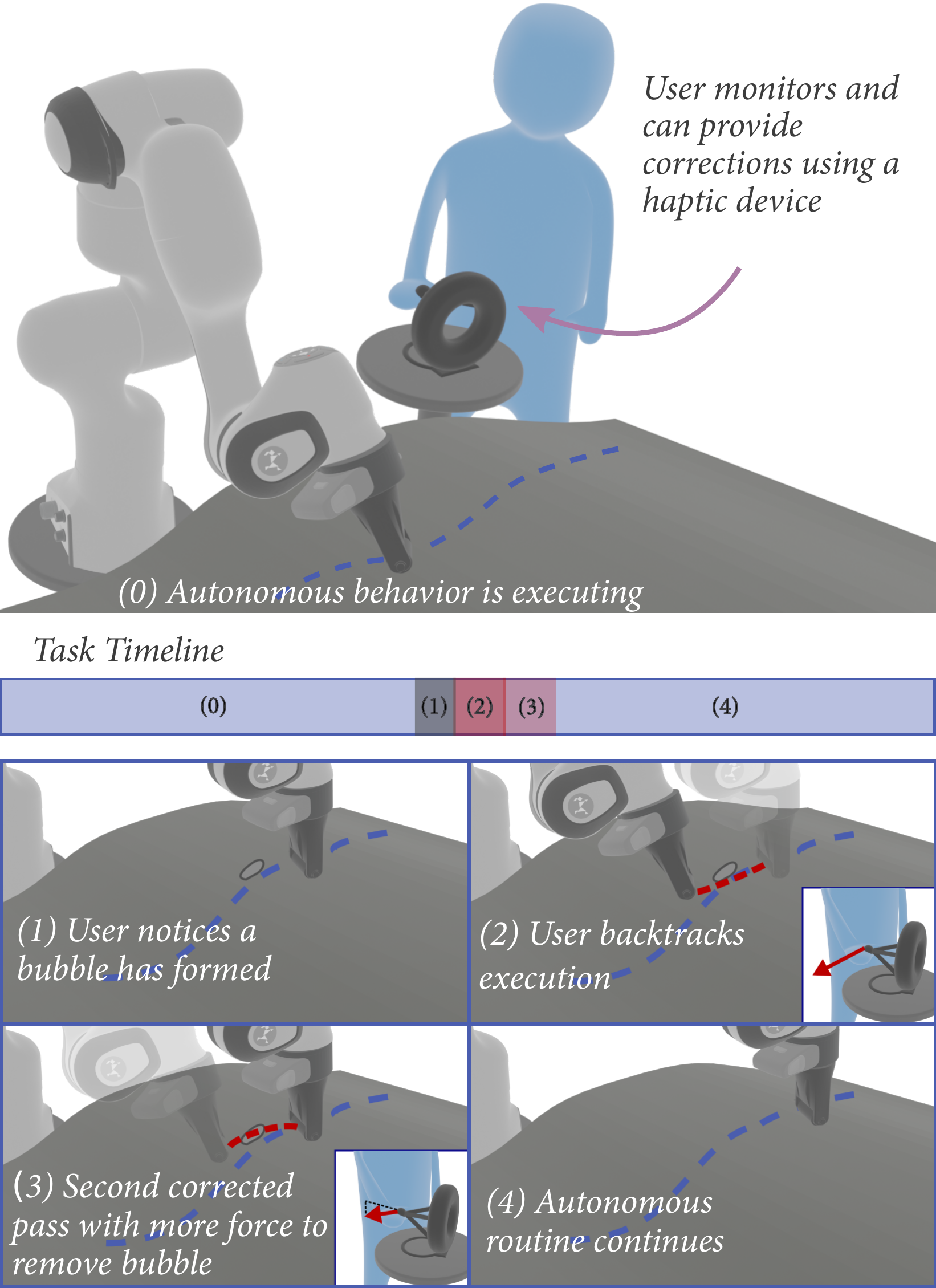}
\caption{Example application of our \emph{corrective shared autonomy} method. A user monitors a robot performing composite layup and provides real-time corrections to kinematic and force robot state variables to counter an unexpected defect. Our method of corrective shared control allows users to address such complicated tasks with a small amount of user input.}
\label{fig:teaser}
\vspace{-15pt}
\end{figure}

We refer to this paradigm of providing intermittent corrections to arbitrary robot state variables as \emph{corrective shared autonomy}. While previous work, described in Section \ref{sec:relatedwork}, has explored kinematic corrections of robotic platforms, our vision is generalizing corrections to any relevant robot state variable required to address the uncertainty of a given task. For example, in Figure \ref{fig:teaser}, a technician uses a three-degree-of-freedom device to make corrections to surface coordinates, normal force, and execution rate to counter intermittent bubbles during a composite layup process. While the generalization of robot state corrections enables application to a wider range of tasks, this flexibility introduces challenges and design considerations to the shared autonomy interface which are discussed throughout this work.

Our shared autonomy method addresses variability in tasks by enabling users to provide real-time corrections to key robot state variables on top of a nominal autonomous task model. Our contributions include (1) generalizing real-time corrections beyond kinematic robot state variables, (2) presenting an implementation of \emph{corrective shared autonomy} focused on tasks involving physical interaction, where users provide corrections to variables such as positions, forces, surface coordinates, and rate of execution, and (3) providing preliminary evidence through a user study that this approach enables users to complete a range of physical tasks situated in aircraft manufacturing.
\section{RELATED WORK}
\label{sec:relatedwork}
In our approach, users provide corrections to key robot state variables in order to address task uncertainty. In contextualizing our work, we refer to \emph{shared control} as methods where the human drives the system behavior with assistance and \emph{shared autonomy} as the broader classification of methods combining human input and autonomous behavior to achieve shared goals \cite{Javdani2018}. We provide a brief review of some guiding seminal work in the shared control field and two previous research threads in shared autonomy, \emph{role arbitration} and \emph{real-time corrections}, that serve as inspiration for our method.

Existing methods in shared control are often derived from adding assistive policies to teleoperation and are well-summarized in the literature \cite{dragan2013}\cite{LoseyReview2018}\cite{Boer2018}. Recent works have determined a variety of ways an assistive policy can be derived. As recent examples, Zeestraten et al. \cite{CalinonRAL2018} encode a shared control policy using a gaussian mixture model (GMM) on a Riemannian manifold via learning from demonstration and Abi-Farraj et al. \cite{abifarraj2020} encode a policy based on grasp poses. Similar to methods both within shared control and beyond, we leverage subspace input as a means to reduce operator workload, both physically and mentally. For example, many shared control policies use a divisible shared control law (e.g.,\cite{Tee2018},\cite{omalley2005}) where the human and robot control complementary subspaces of the task space.

Methods in dynamic role arbitration, where the user and robot exchange leader roles,  allow a robot to function autonomously in the absence of user intervention, but are not well suited for addressing corrections to general robot state variables. Both Medina et al. \cite{Medina2013} and Evrard el al. \cite{Evrard2009} explore the leader-follower paradigm in the context of physically coupled cooperative transport based on measuring disagreement forces between the human and robot. Li et al. \cite{Li2015} similarly propose a scheme for continuous role adaption of a human-robot team where the control arbitration is modeled using a two agent game theoretic framework. While these approaches share motivation to leverage an autonomous behavior to avoid excessive human input and the resulting effects on cognitive load and performance, the approaches are focused on physical human robot interaction, or in other words the user specifies a desired robot state by applying forces to the robot directly. 

While previous work has looked at providing corrections to robot trajectories, the focus has been largely on kinematics and using corrections to inform robot learning rather than providing an effective means to control for task uncertainty. Losey et al. \cite{LoseyDeformations2018} propose an energy-based methodology for a robot to deform its trajectory based on physical corrections applied to the robot. In \cite{LoseyPHRI2020},\cite{LoseyPolicy2017}, this work is extended to use the kinematic corrections to successfully infer parameters of an optimal policy. Nemec et al. \cite{Ude2018} propose a programming by demonstration framework where dynamic movement primitives are refined based on kinematic corrections to the nominal behavior of an impedance-controlled robot. As final examples, Masone et al. \cite{Masone2014} and Cognetti et al. \cite{cognetti2020} explore providing canonical path transformations (e.g., translation, scaling, rotation) to B-Spline paths to affect the trajectory of UAVs and mobile robots respectively.

Our method is contextualized in shared autonomy and designed for environments where the task model has a strong prior. In contrast to previous methods, we focus on generalizing the types of corrections that a user can make from kinematics to any state variable that is pertinent to address the variability of the task (e.g., forces, execution rate).
With respect to policy learning, we believe that adding an element of learning from corrections could serve as a complementary capability in the future.

\begin{figure*}
\centering
\vspace{5pt}
\includegraphics[width=.95\textwidth]{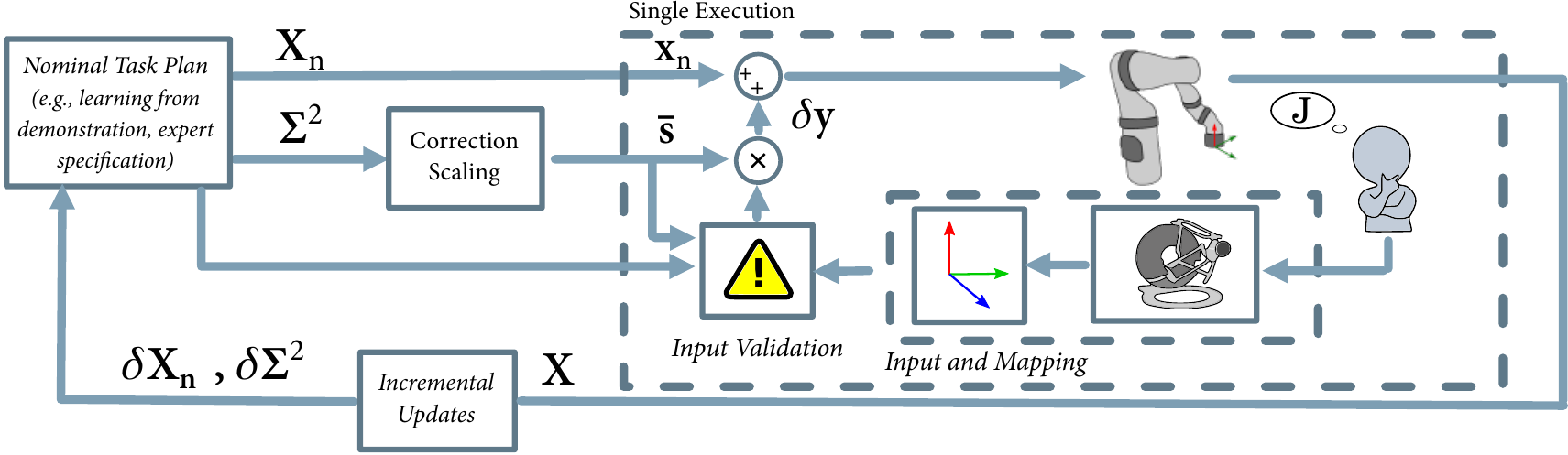}
\caption{Block diagram illustrating the topology and key design decisions related to the \emph{corrective shared autonomy} system. $\textbf{X}_n$ and $\mathbf{\Sigma}^{2}$ are the full nominal behavior and variance respectively. These are used to define the nominal state, $\textbf{x}_n$ and the correction scaling, $\bar{\textbf{s}}$. During execution, the user makes corrections based on an internal, expert cost function, $J$, and provides corrections in a subspace of the state vector. After the execution, the true behavior, $\textbf{X}$, can be used to incrementally update the policy and variance. In this work, we focus on elements within the 'Single Execution' box.}
\label{fig:correctionsblockdiagram}
\end{figure*}

\section{CORRECTIVE SHARED AUTONOMY}
The high level concept of \emph{corrective shared autonomy} can be described via a simple arbitration. Autonomous input from a nominal task model is summed together with relative input provided by the human based on observations of any errors caused by the current robot state. The operator provides these corrections to key robot state variables that may vary over the course of the task execution. Notably, these corrections are often based on implicit feedback (e.g., corrections to force in sanding based on sound or visual results). The final commanded robot state can be determined as:
\begin{align}
        \textbf{x} = \textbf{x}_{n} + \delta \textbf{y},\textbf{x}_{n} \in \mathbb{R}^{m}, \delta \textbf{y} \in S(\mathbb{R}^{m})
\label{eq:arbitration}
\end{align}
where $\textbf{x}_{n}$ is the nominal robot state, $\delta \textbf{y}$ is the user correction, $m$ is the dimension of the robot state, and $S(\mathbb{R}^{m})$ is the subspace where users can provide corrections. A \emph{corrective shared autonomy} system must be appropriately scoped to the types of corrections a user should be able to provide which consequently informs the general system requirements.

\subsection{Types of Corrections}
\label{sec:correctiontypes}
Based on the task, there are a variety of forms a correction might take. For the purposes of our method, we propose they can broadly be classified into two categories: \emph{prior} and \emph{posterior} corrections. A \emph{prior} correction is when the user is able to see that a correction is going to be needed and is able to intervene prior to the issue causing undesired consequences. For example, an operator might notice that a joining operation is not properly aligned and is able to provide the corrections prior to the pieces incorrectly mating. A \emph{posterior} correction is when the user notes that a failure has occurred during the execution and wants to be able to go back to address it. For example, an operator may notice that a spot is missed during a spray-painting operation and desire to recoat with a corrected, lighter spray rate.

The frequency of errors and resulting corrections may also vary depending on a given task. A corrections-based policy is particularly effective when few corrections are needed to the autonomous behavior. We argue that in many tasks, there can be small regions of variability where a user's input is needed. For example, during a drilling and fastener installation operation, a majority of the execution may be able to execute without assistance, including attaching the drill, moving the drill to the drilling location, switching to the fastener jig, grabbing a fastener, and moving back to the location. In this example, it can be determined ahead of time that the variable regions of the task consist of when the drill contacts the surface, drills the hole, and when the fastener is aligned for insertion into the drilled hole.

\subsection{System Requirements}
A \emph{corrective shared autonomy} system requires a variety of design decisions that are illustrated in Figure \ref{fig:correctionsblockdiagram}. First and foremost are choices in representation. This includes both how the nominal behavior is modeled as well as a model of how corrections are layered on top. The nominal behavior must be amenable to the arbitration in Equation \ref{eq:arbitration} (e.g., encoded trajectory). Other behaviors, such as logic-based behaviors (e.g., robot must approach and pass through a certain waypoint), may require additional consideration in allowing corrections. Depending on the type of corrections as described above, there also may be a need to either slow (e.g., \emph{prior} correction) or reverse (e.g., \emph{posterior} correction) the execution which should be considered in the choice of the nominal behavior representation. Furthermore, there is also a consideration of how the nominal model might refine or generalize over time and how the variability of the task maps to specific corrections, both variables and scaling, a user might make during the real time execution. For example, it may be desirable for the scaling to follow the confidence (e.g., variance in demonstrations) such that the ranges of corrections is appropriate and confident regions of robot execution suppress errant corrections. However, if the demonstrations are mistakenly confident or an element of the task changes (e.g., new collision), it may also be desirable to include a user override of the correction scaling to maintain final control. In this work, we focus on elements of the real-time execution during \emph{corrective shared autonomy} and leave the ability to incrementally update the behavior, as well as how to automatically adapt scaling and choose correction variables to future work.

Within the real-time execution, the use of a decoupled input affords many choices with how to provide corrective input. The largest benefit is that a decoupled device trivially allows for corrections to any state variable. This requires that there is appropriate choice of input mapping. For example, if the state variables do not have a clear spatial direction for input (e.g., tool speed), additional visualization may be required to assist the user in mapping their input.
\section{IMPLEMENTATION AND SYSTEM DESIGN}
We designed an instantiation of \emph{corrective shared autonomy} which allows users to provide corrections on tasks involving both position and hybrid control (i.e., position and force modulation) against arbitrary geometries.
We view corrections to hybrid control variables as an important step in the direction of generalized task variable corrections, which covers many tasks in manufacturing involving interaction with the environment. For example, the validation tasks in this paper include fastener insertion, polishing, and composite layup. Users can provide simultaneous corrections to three robot state variables using a three-degree-of-freedom haptic device operating in a position regulation mode, similar to a joystick. In this implementation, the choice of correction variables is chosen to be spatially consistent with the control goals of the robot (e.g., corrections to the three spatial directions during free space motion, corrections to two spatial and one force variable when interacting with a surface).  Designing this system required choices for how to (A) represent the autonomous behavior and the various forms of corrections, (B) allow for proper surface interaction, (C) map user input to corrections on the robot, (D) allow modulation of the execution rate, (E) transition between contact and non-contact segments, (F) validate the user input, and (G) derive the nominal behavior and scale the corrections.
The remainder of this section describes the pertinent design choices of this implementation.
\subsection{Representation}
Providing layered corrections requires a representation for both the nominal autonomous behavior and the layer of real-time input from the user. While any form of open-loop trajectory encoding (e.g., DMP, ProMP, GMM, quintic polynomials) can be used for the nominal autonomous behavior, we choose to represent the nominal trajectory with a sequence of dynamic movement primitives \cite{Ijspeert2013}. One value of such regression methods is that they can encode any arbitrary set of state variables. Dynamic movement primitives (DMPs) in particular offer several benefits for encoding general robot behaviors, including smoothness, the ability to be acquired through learning from demonstration, and temporal coupling that makes augmentation of the execution rate trivial. Additionally, the framework is suited for simple reuse between different behaviors while maintaining stability. We provide a partial review of the dynamic movement framework for context. A dynamic movement primitive for a state vector, $\textbf{x}$, can be represented as:

\begin{equation}
\begin{bmatrix}
   \tau\dot{s} \\
   \tau\dot{\textbf{x}} \\
   \tau\dot{\textbf{z}}
   \end{bmatrix} = \begin{bmatrix}
   -as \\
   \textbf{z} \\
   \alpha(\beta(\textbf{g}-\textbf{x})-\textbf{z})+\textbf{f}(s)
   \end{bmatrix}
\end{equation}
where $s$ is the phase variable of the canonical system, $a$ is the related constant of the canonical system, $\tau$ is the time constant of the second-order dynamical system, $\textbf{g}$ is the goal state, $\textbf{z}$ is the time derivative of the state, $\alpha$ and $\beta$ are positive constants determining the roots of the dynamical system, and $\textbf{f}(s)$ is the nonlinear forcing function defined for a single state as:
\begin{equation}
	f(s) = \left(\sum\limits_{i=1}^{N}\psi_{i}(s)w_{i}\right)/\sum\limits_{i=1}^{N}\psi_{i}(s)
\end{equation}
where $\psi_{i}$ are Gaussian basis functions with corresponding weights, $w_{i}$. Weights are acquired using locally weighted regression (LWR) \cite{CalinonLee19}. An independent DMP is learned for each state variable (e.g, $x$, $y$, $f_{z}$) and a separate set of DMPs is learned for each segment of the trajectory (e.g., free space, hybrid control). For example, a task consisting of approaching a surface, drawing a line, then receding would consist of three sets of DMPs where each set contains $m$ DMPs corresponding to the $m$ state variables.

In this instantiation, orientation can either be prescribed, constant, or inferred depending on the nature of the task. When prescribed (e.g., free space motion), orientation is represented using quaternions. Since the four quaternion entries are mathematically coupled, it is common to use a modified framework, CDMPs \cite{Morimoto2014}, to represent orientation rather than independent DMPs. In this work, the prescribed orientations change sufficiently slowly that a linear interpolation and re-normalization procedure is used instead.

The user is able to add a correction on top of the nominal behavior. A correction to a robot state is represented using a second-order ordinary differential equation:

\begin{equation}
	\ddot{\delta \textbf{y}} + b_{c} \dot{\delta \textbf{y}} + k_{c}{\delta \textbf{y}} = \textbf{u}
\end{equation}
where ${\delta \textbf{y}}$ is the correction output, $b_{c}$ and $k_{c}$ are damping and stiffness parameters for the correction dynamical system, and $\textbf{u}$ is the user input. The output value, $\delta \textbf{y}$, is computed using numerical integration similar to the DMPs. The output of the dynamical system has a bounded maximum value ($k_{c}/\textbf{u}$, entries of $\textbf{u}$ are unit-bounded). Thus, the system output is linearly scaled to give the appropriate final correction scaling. The designer can choose $k_{c}$ based on desired correction speed (in this work, twice the speed of the nominal system based on experimental tuning). In general, this value should balance an achievable correction speed for the robot (e.g., velocity limits) while being sufficiently responsive to the users commands. This dynamical system also effectively filters the user input command which can be useful to remove user jitter. In our implementation, the damping and stiffness are equivalent for each correction state and set to be critically damped. Depending on the correction variable, it may be desirable to have a different time constant.

\subsection{Surface Interaction}
In order to allow for corrections to surface coordinates, the \emph{corrective shared autonomy} system requires a surface model where the surface points and normals can be extracted for points within the local neighborhood of the nominal autonomous behavior. We choose one common tool, B-Spline Surfaces \cite{Mortenson1990}, where the surface position, $\textbf{r}$, can be determined according to:
\begin{equation}
    \textbf{r}(u,v)=\sum\limits_{i=0}^{m}\sum\limits_{j=0}^{n}N_{i,k}(u)N_{j,l}(v)\textbf{p}_{ij}
\end{equation}
where $N$ are the surface knot vectors and $\textbf{p}_{ij}$ are the surface control points, which are determined analytically from the geometry when possible, or in other cases using the method described in \cite{Kai1998}.
When interacting with a surface, the admissible orientation is determined by constraints on the specific end-effector. In most cases, it is desired to have the tool perpendicular to the surface, however, this leaves the orientation underdetermined. For certain end-effectors (e.g., polishing, sanding), any rotation is permissible and the remaining direction can be chosen to be static (i.e., a specific surface direction). Other end-effectors, such as a roller for composite layup, require that the end-effector (e.g. roller) is consistent with the direction of motion.

\subsection{Input mapping}
To reduce the number of mental rotations for the user, we map the input device to be spatially consistent with the orientation of the device relative to the robot. When controlling on the surface, we assume the surfaces are sufficiently homeomorphic to a plane and compute one additional static rotation matrix to define the correction frame (i.e., which input directions map to positions vs forces against the surface). This rotation matrix is determined by finding the closest fit plane to the control points:
\begin{equation}
    \argmin_{\tilde{\textbf{w}},d}{{\sum\limits_{i}\sum\limits_{j}\left( {{e}^{{\tilde{\mathbf{w}}}}}{{\left[ 0\ \ 0\ \ d \right]}^{T}}-\textbf{P}_{ij} \right)}^{T}}{{e}^{{\tilde{\mathbf{w}}}}}{{\left[ 0\ \ 0\ \ 1 \right]}^{T}}
\end{equation}
where $\tilde{\textbf{w}}= [w_x\ w_y\ 0]^{T}$ and $d$ are the exponential coordinates and distance of the plane from the origin respectively. The resulting nonlinear optimization is solved using Nelder-Mead \cite{neldermead}. From the best fit plane, the control points are projected into the plane to align the control point directions back into the global frame. The result is a static rotation matrix that can be used to map user input into the surface coordinate frame. In the future, we plan to explore the impact of different control 
frames (e.g., camera frame for remote presence, control for higher degrees of curvature).

\subsection{Execution Rate}
For many corrections, it may be desirable to slow the execution (e.g., approaching a sensitive area or issue) or potentially to backtrace in order to perform a posterior correction (e.g., go back and retrace a section that was missed). To address these design criteria, we allow the user to both slow and reverse the trajectory within a local region (e.g., the current dynamic movement primitive segment). To allow for variation of the execution rate without the need for an additional input device, we use a basic heuristic where the execution speed is based on the direction relative to the current velocity of the system similar to the heuristic proposed by Ude et al \cite{Ude2018}.
\begin{equation}
    \tau=\left\{
\begin{array}{ll}
      (1+\gamma(\textbf{v}_n \cdot \textbf{f}_d(s)))^{-1} & (\textbf{v}_n \cdot \textbf{f}_d(s))\leq 0 \\
      1 & (\textbf{v}_n \cdot \textbf{f}_d(s))>0 
\end{array} 
\right.
\end{equation}
where $\textbf{f}_{d}(t)$ is the correction direction (scaled to a maximum of one), $\textbf{v}_n$ is the current normalized velocity, and $\gamma$ is a flexible parameter that specifies how much the trajectory can be slowed or reversed. If $0\leq\gamma\leq1$, the system only allows slowing the execution. If $\gamma>1$, the time constant can flip sign allowing backtracking.

One challenge introduced by backtracking is that the roots of the dynamic movement primitive system become unstable when $\tau$ becomes negative. To address this, we learn forwards and backwards systems for each DMP and switch depending on the sign of the time constant. In order to preserve continuity of the heuristic (i.e., pushing against the current direction of motion continues to backtrack even once the system changes its direction of motion), it is necessary to add some additional triggers to the velocity heuristic that look for when the direction of motion inverts as the forwards and backwards DMPs transition. In the future, we are interested in looking at other systems that might more easily represent both directions of behavior such as logistic differential equations \cite{Savarimuthu2019}.

In cases where the velocity is frequently changing in cardinality within a small neighboring window (e.g., circular motions), this heuristic might not be effective and dedicating an input degree of freedom to specifically modifying the execution rate might be more effective. Additionally, our velocity heuristic allows for a maximum time constant of one, or in other words, the rate of execution cannot be increased. We assume that the nominal behavior is designed to run near the maximum speed for the robotic platform, however, in other cases, it may be desirable to allow the execution rate to increase by modifying the heuristic.

\subsection{Transitions between behavior}
One challenge introduced by real-time corrections for segmented behavior (e.g. hybrid control) is that corrections at the end of each section may influence continuity into the next section. For example, if a user corrects the trajectory while approaching a surface to a different location on the surface, it would be desirable to update the starting location of the next DMP on the surface.

In this implementation, if the correctable variables are the same between sections, the starting points of the next DMP can simply be updated. If the transition is from free-space to a surface, it is necessary to optimize to find a new starting point in terms of the parameterized surface coordinates. Finally, transitioning from the surface back to free-space requires setting the new starting point based on the surface point of the final parameterized coordinates. If the variables are not continued (e.g., force to position, position to force), the corrections are not transitioned.

\subsection{User Input Validation}
Users provide a layer of corrections on top of the nominal autonomous behavior. There are a few cases where it may be desirable to saturate user input to prevent unintentional harm. For example, in carefully calibrated processes (e.g., tool changes), the correction scaling can converge towards zero as a means to disallow errant input (e.g., bumping the controls). In this instantiation, the user is unable to provide input that would move the robot off the edge of the surface during hybrid control, and when leaving or moving towards the surface, the user's corrections are diminished if they would cause collision with the surface. 

\subsection{Nominal Behavior, Learning, and Scaling}
In this work, both the nominal autonomous behavior and the scaling for each of the corrections are determined manually based on experimental tuning and expert input. In the future, we plan to explore ways that the relevant variables and scaling for corrections can be inferred automatically. Similarly, this work uses a three-degree-of-freedom input which allows for corrections to three variables. In future work, we plan to use demonstrations of a task to infer how many correction variables are needed.
\section{EXPERIMENTAL EVALUATION} \label{sec:experiments}
\begin{figure}[!h]
\centering
\includegraphics[width=3.2in]{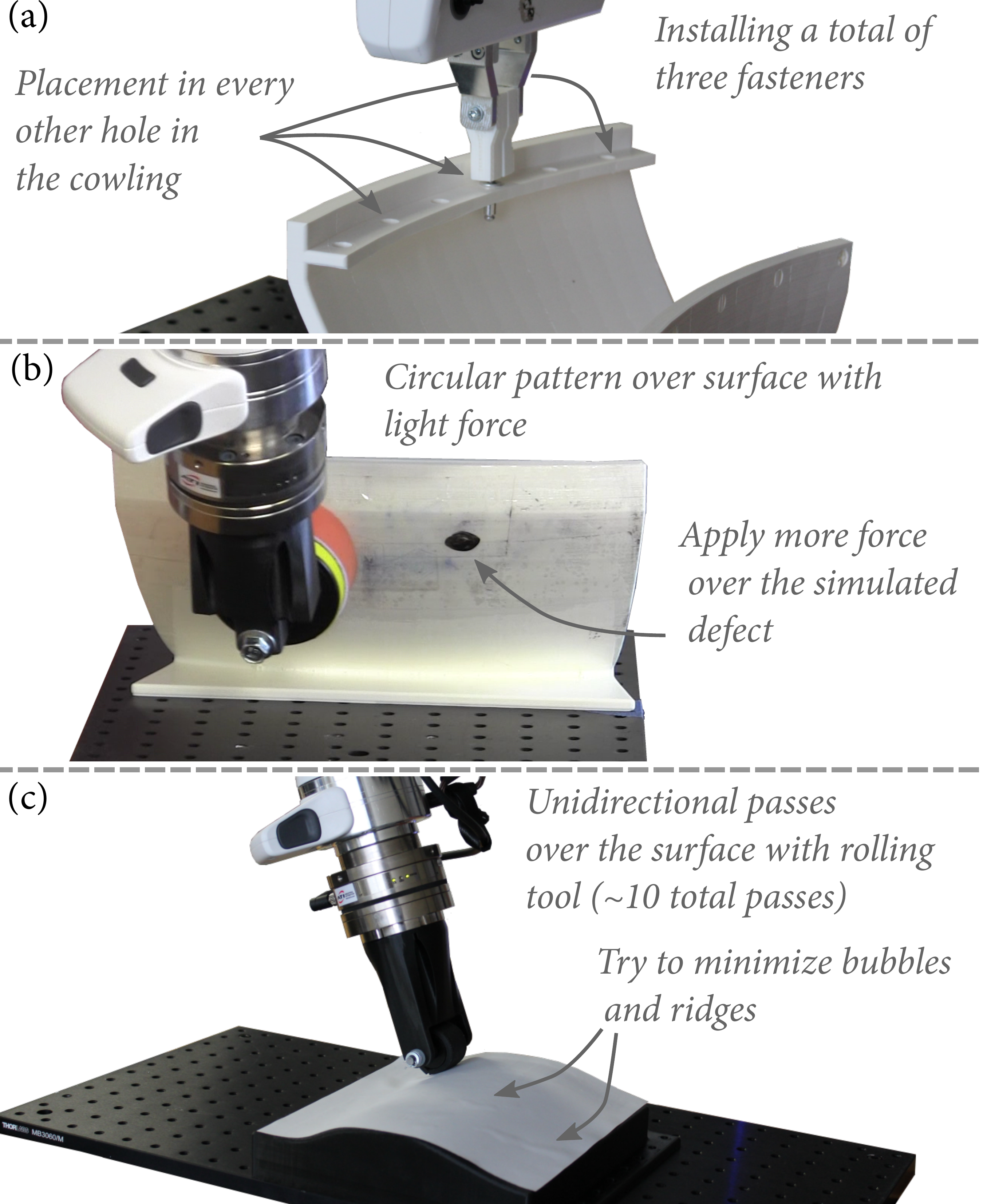}
\caption{ Three experimental tasks used in the study: (a) inserting fasteners into the top ridge of the cowling, (b) polishing the exterior surface of the cowling, and (c) layup of a vinyl onto a shape resembling a wing curvature.}
\label{fig:exptasks}
\end{figure}

We performed a preliminary system validation study involving six participants (four male, two female) with ages 20--27 ($M=21.8$, $SD=2.5$) recruited from a university campus. The procedure was administered under a protocol reviewed and approved by the Institutional Review Board (IRB) of the UW–-Madison. The study followed a within-subjects design where after providing written informed consent, participants completed three manufacturing tasks (shown in Figure \ref{fig:exptasks}) under three conditions presented in a counter-balanced order, for a total of nine evaluation trials. Before each evaluation trial, participants performed one practice trial. Participants were shown a video of the task being done manually before practicing. These tasks were designed based on manual tasks from an aviation manufacturing context that involve some level of variability. In all conditions, the robot was mounted on one end of a table and participants were stationed on the other end to both prevent the robot occluding the view and keep participants outside the robot reach during operation (approximately 1 meter from the task piece). The tasks are detailed in a later section.

\subsection{System Details}
All experiments were performed using a Franka Emika Panda collaborative robot equipped with an ATI Axia80 6-axis force-torque sensor and custom end-effectors suited for each task. The robot was commanded in joint velocity using pseudo-inverse based inverse kinematics, and the hybrid control was implemented using an admittance model (i.e., reading forces, commanding velocity with a low proportional gain). To better facilitate contact with the environment, the robot was operated in joint impedance mode (400 N/rad). Users provided input through a Force Dimension Omega 3 haptic device that was operated in a zero displacement mode enforced by a critically damped proportional-derivative (PD) control law.

\subsection{Conditions}
We compared against both manual completion of the task as well as a common shared control policy for teleoperation systems. Rather than one of the physical human-robot interaction methods described in Section \ref{sec:relatedwork}, we chose a teleoperation-based shared control method to compare robotic methods using the same input device. 

\begin{figure*}[!b]
\centering
\includegraphics[width=7in]{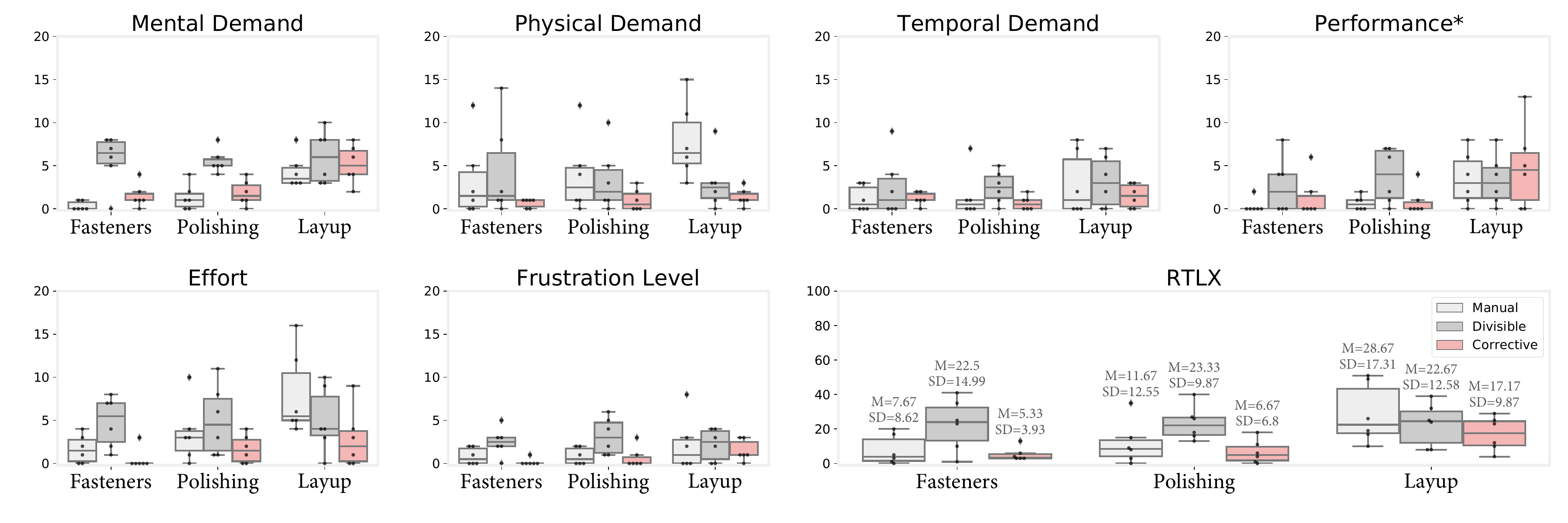}
\vspace{-15pt}
\caption{NASA TLX metrics reported for each task including the Raw Task Load Index (RTLX) comparing the manual, divisible shared control and \emph{corrective shared autonomy} methods. * Scale is flipped such that lower values are better.}
\label{fig:nasatlxresults}
\end{figure*}

\subsubsection{Manual Task}
Since variants of these tasks are commonly done manually in industry, we used manual as our performance baseline. In this condition, participants performed the same operations by hand, either using a single motion-capture-tracked hand or tracked tool (e.g., roller, polishing pad) similar to the end-effectors in Figure \ref{fig:exptasks}. We did not design the tasks to be physically taxing but rather focused on using the manual data to benchmark desirable performance.

\subsubsection{Divisible Shared Control}
We compared against a basic reduced input shared control policy where the user controlled all position degrees of freedom, and the robot automatically controlled the complementary orientation based on the closest point on the surface and the corresponding normal vector. This method was selected to offer some shared assistance and bilateral feedback without introducing the issue of coupling guidance and interaction force feedback \cite{omalley2012}. This method used a static four-to-one (robot-to-input) workspace mapping with clutching functionality for larger motions. The method was bilateral with a 60-percent force scaling and active damping to maintain margin for the bilateral stability. These values were tuned to give the best performance for the target tasks, noting that it is challenging to render stiff interactions with an impedance-based master device.

\subsubsection{Corrective Shared Autonomy}
For each task, a nominal trajectory was constructed for the corrections-based method by manually selecting waypoints and encoded using DMPs. The one-time learning of the behaviors ranged from 9 to 133 seconds. Each nominal execution was designed to include at least one error in the absence of user input (i.e., simulating a region of variability). The imperfections were added manually in deterministic locations to assure proper control of injected errors. Three-degree-of-freedom input was provided using the Force Dimension. In this implementation, the device provided no haptic feedback (e.g., force reflection, haptic cues). Users provided corrections to all relevant state variables based on visual and auditory feedback of the task. Details are provided in each task below.

\subsection{Study Tasks}
\subsubsection{Task 1. Rivet Fastener Insertion}
\label{sec:fastenertask}
Users had to place three 6.35 mm (1/4") rivets into designated fastener locations on a prototype section of an inlet cowling. The three fasteners were initially located in a holder to the right of the cowling. The nominal autonomous behavior was designed such that grabbing the rivets from the holder was always successful (assuming the holder was local to the robot), but during the placement, two rivets had 3 mm of alignment error, simulating registration error between the robot and workpiece.

\subsubsection{Task 2. Exterior polishing of an Engine Cowling}
\label{sec:polishingtask}
Users were instructed to polish the top of the exterior of the engine inlet cowling, including an imperfection (black marker) that required more force to buff out. Users were instructed to apply increased force over this region. In the corrective method, users generally modulated force based on auditory cues (e.g., squeaking). The nominal autonomous robot behavior was designed to polish with a consistent low force, requiring user intervention for the marked region.

\subsubsection{Task 3. Composite Layup of a Curved Surface}
\label{sec:compositetask}
We simulated composite layup manufacturing by constructing a surface resembling the typical curvature of a jet airfoil. The user was instructed to apply a vinyl fabric with minimal wrinkles and bubbles (i.e., consistent adherence to the surface) using a rolling pin tool similar to a hand tool a factory worker might use. The nominal autonomous behavior was constructed to perform ten horizontal passes across the surface, including one pass that was intentionally misaligned to cause a crease to form without user intervention. The base execution speed was constant across users and intentionally slowed below the robot's capabilities to give users time to spot issues. In general, it may be desirable to set the speed based on a particular user's comfortable reaction time. 

\subsection{Measures \& Analysis}
After each recorded trial, participants completed the NASA TLX questionnaire \cite{NASATLX}. Additionally, using motion capture and the Force Dimension, user input was measured to determine the user input time (assessing whether the designed idle time of the correction-based method manifested in practice):
\begin{equation}
    t_{input} =\int_{0}^{t_f}\mathbb{1}_{m(t)}dt
\end{equation}
\begin{equation}
    m(t) = \left\{
\begin{array}{ll}
      ||\textbf{x}-\textbf{x}_{idle}||>d & corrective \\
      ||\textbf{v}||>v_{\alpha} & else \\
\end{array} 
\right.
\end{equation}
where $m(t)$ is a boolean function for whether the participant is providing input; $\textbf{x}$ is the haptic device location; $\textbf{x}_{idle}$ is the location of the haptic device when no input is applied; $d$ is the distance threshold that is considered user input ($d=5$ mm); $v$ is the user's filtered velocity (e.g., hand or haptic input); $r_{total}$ is the total distance traveled; $p_{idle}$ is the user idle percentage; and $v_{\alpha}$ is the threshold used to define whether a user was idle (in this experiment, $v_{\alpha}=0.01 m/s$). The corrective method is considered as providing input when the haptic device is displaced from the neutral position (the device can be stationary, but providing a fixed corrective input), whereas for the other methods it is whether the input is moving.
Given the preliminary nature and small sample size of the experiment, we report only descriptive statistics.

\subsection{Results}
Participants were generally able to complete all tasks with all methods with the exception of one dropped rivet in both the corrective and divisible methods. Participants achieved similar levels of quality for the polishing and layup tasks across the three methods, though there were observable differences in how participants perceived task-method combinations. Figure \ref{fig:nasatlxresults} summarizes the NASA TLX results. The \emph{corrective shared autonomy} tended to perform favorably in some NASA TLX categories, particularly physical demand and effort. In the polishing and insertion tasks, our method tended to result in a lower workload than the divisible method. Our method tended to be at least comparable in workload to the other conditions for all tasks.

\begin{table}
\vspace{5pt}
  \caption{Input and total time for each task. Values denote mean (SD).}
  \label{table:quantitative}
  \setlength\tabcolsep{4pt}
  \begin{tabular}{p{1.0cm}lp{1.33cm}p{1.33cm}p{1.40cm}}
    \toprule
    Task & Metric & Manual & Divisible & Corrective \\    
    \midrule
    \multirow{3}{*}{Fasteners} & Input Time (s) & 7.9 (2.3) & 34.9 (6.5) & 9.8 (3.0) \\
 & Total Time (s) & 8.5 (2.9) & 64.0 (19.4) & 44.3 (1.3) \\     
    \midrule
\multirow{3}{*}{Polishing} & Input Time (s) & 7.7 (2.6) & 21.5 (7.6) & 6.3 (1.7) \\
 & Total time (s) & 7.7 (2.6) & 24.1 (9.7) & 12.3 (0.8) \\    
    \midrule
\multirow{3}{*}{Layup} & Input Time (s) & 45.9 (22.8) & 46.5 (12.1) & 14.7 (11.6) \\
 & Total Time (s) & 47.2 (25.0) & 82.7 (47.7) & 118.6 (13.6) \\
\bottomrule
  \end{tabular}
\end{table}

The human input and total times are reported in Table \ref{table:quantitative}. For all tasks, the total input time and total time tended to be lower for the corrective method than the divisible method. The input times for the riveting and polishing tasks tended to be similar the corrective and manual methods, however, for the longer-duration layup task, the input time for the corrective method tended to be lower.

\section{GENERAL DISCUSSION AND CONCLUSION}
The low human input time (i.e., increased idle time) of the corrective method affords additional operator productivity. Users can potentially complete a secondary task (e.g., sorting, monitoring) or provide corrections to a second robotic setup during these idle times. In the future, we will explore such opportunities including leveraging regions of intermittent variability fault detection to allow intervention across multiple robotic setups.

Our method assumes the task model has a sufficiently strong prior that can be used to construct a nominal autonomous behavior and that the user is aware of the plan of the nominal autonomous execution. The efficacy of our method is also increased for tasks where the regions of variability represent smaller percentages of the overall execution. Additionally, the efficacy of the method is dictated by the quality of the nominal model. If a nominal model is poor and the user must provide frequent corrections, opportunities to leverage the decreased human input become less valuable.
In this work, we focused the instantiation and discussion around elements of the execution. In the future, we plan to expand this work to infer what variables and scaling are needed for a given task. This includes using elements of confidence (e.g., learning from demonstration) in addition to safeties and conservative overrides. We also plan to investigate the space of design choices (e.g., representation, input mapping, saturation and validation, feedback) and how they influence the user's perception of the corrections-based method.

In this paper, we presented \emph{corrective shared autonomy}, a method to leverage user input to address uncertainty in robot tasks by targeting corrections to task-specific variables. We provided details of a specific implementation focused on tasks involving physical interaction and hybrid control. We conducted a user study that showed the potential of such a method for tasks involving physical interaction.

\bibliographystyle{IEEEtran}
\bibliography{references.bib}
\end{document}